\documentclass[pdflatex,sn-mathphys-num]{sn-jnl}

\usepackage[utf8]{inputenc}
\usepackage[T1]{fontenc}
\usepackage{listings}
\usepackage[mathscr]{eucal}
\usepackage{caption}
\usepackage{amsmath,amssymb,amsfonts}
\usepackage{graphicx}
\usepackage{bm}
\usepackage{float}
\usepackage{booktabs}
\usepackage{multirow}
\usepackage{array,tabularx}
\usepackage[table]{xcolor}

\usepackage[table]{xcolor}
\definecolor{E3D5CA}{HTML}{E3D5CA}
\definecolor{EDEDE9}{HTML}{EDEDE9}

\usepackage{algorithm}
\usepackage{algpseudocode}

\usepackage{url}
\usepackage{hyperref}
\usepackage[table]{xcolor} 

\definecolor{EDEDE9}{HTML}{EDEDE9}

\usepackage{caption}
\usepackage{subcaption}

\usepackage[title]{appendix}
\theoremstyle{thmstyleone}%
\theoremstyle{thmstyletwo}%

\theoremstyle{thmstylethree}%

\makeatletter
\@ifpackageloaded{caption}{%
  \captionsetup{format=plain,labelformat=simple,labelsep=period}
}{}
\makeatother
\begin{document}

\title{Interpretable Machine Learning Model for Early Prediction of Acute Kidney Injury in Critically Ill Patients with Cirrhosis: A Retrospective Study}


\author[1]{\fnm{Li} \sur{Sun}}\email{lsun4765@usc.edu}
\author[1]{\fnm{Shuheng} \sur{Chen}}\email{shuhengc@usc.edu}
\author[1]{\fnm{Junyi} \sur{Fan}}\email{junyifan@usc.edu}
\author[1]{\fnm{Yong} \sur{Si}}\email{yongsi@usc.edu}
\author[1]{\fnm{Minoo} \sur{Ahmadi}}\email{minooahm@usc.edu}
\author[2]{\fnm{Elham} \sur{Pishgar}}\email{dr.elhampishgar@gmail.com}
\author[3]{\fnm{Kamiar} \sur{Alaei}}\email{kamiar.alaei@csulb.edu}
\author*[1]{\fnm{Maryam} \sur{Pishgar}}\email{pishgar@usc.edu}

\affil*[1]{\orgdiv{Department of Industrial and Systems Engineering}, \orgname{University of Southern California}, \orgaddress{ \city{Los Angeles},  \state{CA}, \country{United States}}}
\affil[2]{\orgdiv{Colorectal Research Center}, \orgname{Iran University of Medical Sciences}, \orgaddress{\city{Tehran}, \country{Iran}}}
\affil[3]{\orgdiv{Department of Health Science}, \orgname{California State University, Long Beach}, \orgaddress{ \city{Long Beach}, \state{CA}, \country{United States}}}

\abstract{
\textbf{Background:} Cirrhosis is a progressive, end-stage liver disease associated with profound systemic complications and high mortality. Among these, acute kidney injury (AKI) is one of the most frequent and devastating, occurring in up to 50\% of hospitalized cirrhotic patients and markedly worsening prognosis. AKI in cirrhosis is driven by complex hemodynamic, inflammatory, and metabolic disturbances, and its early detection is crucial for timely intervention. However, existing predictive tools often lack accuracy, interpretability, and alignment with real-world intensive care unit (ICU) workflows. This study aimed to develop an interpretable, high-performing machine learning model for early AKI prediction in critically ill patients with cirrhosis.

\medskip

\textbf{Methods:} We conducted a retrospective analysis using the Medical Information Mart for Intensive Care IV (MIMIC-IV, v2.2) database, identifying 1240 adult ICU patients with a diagnosis of cirrhosis and excluding those with ICU stays $<$48 hours or missing key clinical data. Routinely available laboratory and physiological variables from the first 48 hours of ICU admission were extracted. A machine learning pipeline incorporating structured preprocessing, missingness filtering, LASSO feature selection, and SMOTE class balancing was applied. Six algorithms—LightGBM, CatBoost, XGBoost, logistic regression, naïve Bayes, and neural networks—were trained and evaluated using AUROC, accuracy, F1-score, sensitivity, specificity, and predictive values.

\medskip

\textbf{Results:}LightGBM achieved the best performance (AUROC: 0.808, 95\% CI: 0.741–0.856; accuracy: 0.704; NPV: 0.911). Feature contribution analysis identified prolonged partial thromboplastin time, absence of outside-facility 20G placement, low pH, and altered pO$_2$ as the most influential predictors, aligning with known cirrhosis–AKI pathophysiology and offering direct targets for clinical intervention.

\medskip

\textbf{Conclusion:} The proposed LightGBM-based model enables accurate early AKI risk stratification in ICU patients with cirrhosis using only readily available clinical variables. Its high negative predictive value supports safe de-escalation for low-risk patients, while interpretability facilitates clinician trust and targeted prevention strategies. External multi-center validation and integration into electronic health record systems will be essential next steps toward clinical implementation.
}



\maketitle

\section{Introduction}
Acute kidney injury (AKI) is a frequent and life-threatening complication in cirrhotic patients, particularly among those admitted to intensive care units (ICUs). Recent epidemiological studies suggest that the incidence of AKI in cirrhosis has increased substantially, affecting approximately 20–50\% of hospitalized patients with cirrhosis \cite{gomes2019clinical, tariq2020incidence}, and up to 60\% of those requiring ICU-level care \cite{russ2015acute}. This trend is concerning given that AKI confers a 3–7-fold increased mortality risk among cirrhotic patients \cite{lekakis2025mortality, piano2018epidemiology}. Meta-analyses indicate that in-hospital mortality for cirrhotic patients with AKI approaches 50\%, with 30-day and 90-day survival dropping significantly compared to non-AKI counterparts \cite{angeli2015diagnosis, ning2022impact}.

The pathogenesis of AKI in cirrhosis is multifactorial. Hemodynamic alterations due to portal hypertension lead to splanchnic vasodilation, effective hypovolemia, and decreased renal perfusion \cite{moreau2013acute, de2017acute}. These changes can evolve into ischemic tubular injury or hepatorenal syndrome (HRS), a severe subtype of functional renal failure specific to cirrhosis \cite{lee2021current, nayak2013management}. HRS carries a dire prognosis, with untreated Type 1 HRS exhibiting a median survival under 2 weeks \cite{gines2003hepatorenal}. Moreover, recent studies have shown that patients with ATN and HRS-AKI demonstrate comparably high mortality, underscoring the poor outcomes across AKI subtypes in this population \cite{chung2010diagnostic, kadry2015neutrophil}.

Diagnosis of AKI in cirrhosis remains difficult. Serum creatinine (sCr), the primary biomarker for AKI detection, is often unreliable in this population due to reduced hepatic creatine synthesis, low muscle mass, and interference from elevated bilirubin levels \cite{leelahavanichkul2014comparison, wong2017acute}. These factors result in sCr underestimating true renal dysfunction, thereby delaying recognition. The International Club of Ascites (ICA) and Kidney Disease: Improving Global Outcomes (KDIGO) criteria emphasize relative sCr changes (e.g., $\geq$0.3 mg/dL within 48h), improving sensitivity for AKI detection \cite{levin2013kidney}, but they remain imperfect. While novel biomarkers such as NGAL, IL-18, and cystatin C offer promise in distinguishing structural injury from functional decline \cite{sirota2013urine, saha2022evaluation}, they have not yet achieved widespread clinical adoption.

Cirrhotic patients are also vulnerable to rapid decompensation triggered by infection, sepsis, bleeding, and large-volume paracentesis—all common AKI precipitants in this population \cite{garcia2008acute, belcher2013association}. Systemic inflammation and bacterial translocation further contribute to renal vasoconstriction and tubular damage \cite{alexopoulou2017bacterial}. While risk factors such as high MELD scores, refractory ascites, and spontaneous bacterial peritonitis are well established, clinicians currently lack robust, real-time tools to anticipate AKI onset before irreversible damage occurs \cite{gameiro2018prediction}.

Several prognostic models and machine learning (ML) methods have been developed to predict AKI in general ICU populations \cite{yue2022machine, koyner2018development}, but few focus specifically on cirrhosis or are adapted to its pathophysiology. Existing models either lack interpretability, are based on limited features, or fail to address clinical usability. Moreover, most studies rely solely on creatinine levels, ignoring the complex interactions among liver dysfunction, hemodynamics, and systemic inflammation. There is a pressing need for interpretable, high-performance models tailored to cirrhotic ICU patients, especially during the early stages of care when intervention is most effective.

This study makes several notable contributions to the prediction of acute kidney injury in critically ill patients with cirrhosis: 

1. We established a robust data processing and feature selection framework—combining systematic missingness filtering, LASSO regularization, and expert clinical review—that not only improved model stability and generalizability but also identified physiologically plausible, potentially modifiable predictors (e.g., prolonged PTT, metabolic acidosis, altered oxygenation) directly relevant to cirrhosis–AKI pathophysiology.

2. Development of a high-performing, interpretable prediction model. Using routinely available ICU data from the first 48 hours of admission, we designed a robust machine learning pipeline—combining structured preprocessing, two-stage feature selection, and class balancing—that achieved strong predictive performance (LightGBM AUROC: 0.808; NPV: 0.911), enabling both accurate identification of high-risk patients and safe de-escalation for low-risk patients.

3. Identification of clinically relevant and actionable predictors. Feature contribution analysis, SHAP, LAE revealed physiologically plausible, modifiable factors—including prolonged partial thromboplastin time, absence of outside-facility 20G placement, metabolic acidosis, and altered oxygenation—that align with known cirrhosis–AKI pathophysiology. These predictors offer direct clinical targets for early intervention, such as correction of coagulopathy, optimization of acid–base balance, and maintenance of adequate oxygen delivery.

4. Advancing the field beyond existing models. In contrast to prior cirrhosis-AKI risk tools—often based on static scores or single-timepoint creatinine—our approach incorporates dynamic physiologic variables, interpretable machine learning methods, and a workflow aligned with ICU clinical operations, offering improved applicability to bedside decision-making.

5. Bridging clinical integration and establishing a foundation for future multi-center deployment. Beyond statistical performance, the framework is designed for operationalization within electronic health record systems, allowing automated, real-time risk stratification. This enables timely nephroprotective measures, optimized ICU resource allocation, and enhanced shared decision-making.Although current results are based on single-center retrospective data, the methodological framework supports adaptation to diverse datasets and prospective integration, paving the way for broader clinical adoption once external validation is completed.

\section{Data Source and study design}

This retrospective analysis utilized MIMIC-IV (v2.2) \cite{johnson2020mimic}, a publicly available, de-identified critical care database developed by the Massachusetts Institute of Technology in collaboration with Beth Israel Deaconess Medical Center. Spanning ICU admissions from 2008 to 2019, the database includes detailed, time-stamped records on patient demographics, physiological measurements, laboratory results, medications, and procedures. Using this resource, we designed a binary classification pipeline to estimate the risk of acute kidney injury (AKI) in critically ill patients with cirrhosis based solely on clinical data collected during the early ICU stay. Our end-to-end framework—comprising data cleaning, feature selection, model training, validation, and interpretation—was intentionally structured to ensure clinical validity and facilitate integration into intensive care settings.

To enable timely and interpretable prediction of AKI in cirrhotic ICU patients, we implemented a modular machine learning framework. This pipeline was designed to align with real-world ICU workflows by using only routinely available clinical variables from the initial phase of ICU admission, combined with domain-informed feature engineering, robust model development, and transparent evaluation. The full procedure is summarized below in pseudocode format in 
algorithm \ref{alg:aki_risk_cirrhosis}.
\begin{algorithm}[H]
\caption{\textbf{ML Pipeline for Predicting AKI in ICU Patients with Cirrhosis}}
\label{alg:aki_risk_cirrhosis}
\begin{algorithmic}[1]
\Require MIMIC-IV ICU data with cirrhosis diagnosis
\Ensure Binary prediction: AKI occurrence based on established diagnostic criteria

\State \textbf{Step 1: Cohort Construction}
\State Identify ICU patients with cirrhosis using ICD codes (K74\%, K70\%)  
\State Apply criteria: age 18--80, first ICU stay, ICU stay $\geq$48h  
\State Exclude patients with metastatic cancer or missing key clinical data  

\State \textbf{Step 2: Data Preprocessing}
\State Extract relevant variables from the first 48h of ICU admission  
\State Impute continuous variables via KNN  
\State Encode categorical variables as 0/1 or assume absence  
\State Standardize continuous variables using z-score normalization  

\State \textbf{Step 3: Feature Selection}
\State Remove features with $>$20\% missingness across cohort  
\State Apply LASSO logistic regression to remaining features  
\State Select non-zero coefficient features as final predictors  
\State Review selected features with clinical experts for plausibility  

\State \textbf{Step 4: Class Imbalance Handling}
\State Apply SMOTE within training folds  

\State \textbf{Step 5: Model Development}
\State Split data using stratified 70/30 train-test split  
\ForAll{models $\in$ \{lightgbm, catboost, xgboost, logistic regression, naïve bayes, neural net\}}
    \State Tune hyperparameters via 5-fold cross-validation  
    \State Record validation performance  
\EndFor

\State \textbf{Step 6: Model Evaluation}
\State Evaluate models with AUROC, accuracy, sensitivity, specificity, F1, PPV, NPV  

\State \textbf{Step 7: Statistical Validation}
\State Compare cohorts using t-tests on baseline variables  
\State Perform ablation study to quantify feature contributions  

\State \textbf{Step 8: Model Interpretation}
\State Generate SHAP explanations for global and local predictions  
\State Plot ALE curves to explore nonlinear feature effects  
\end{algorithmic}
\end{algorithm}

\subsection{Study Population}
We retrospectively identified a cohort of ICU patients with cirrhosis from the MIMIC-IV database to investigate the risk of creatinine elevation associated with hepatic dysfunction. A structured screening process was implemented to ensure clinical relevance and minimize potential confounding.

From 65,366 ICU stays, we first restricted the cohort to patients with an ICU length of stay of at least 48 hours (n = 35,794), to ensure adequate observation time for detecting renal changes during critical illness. To eliminate repeated measures and focus on the initial phase of critical care, only the first ICU admission per patient was retained.

We then excluded patients younger than 18 or older than 80 years (final n = 29,705) to reduce physiologic heterogeneity related to pediatric renal immaturity and age-associated comorbidities. Cirrhosis was identified using ICD codes beginning with K74 or K70, capturing both alcoholic and non-alcoholic etiologies, which resulted in 1,552 ICU stays. 

To avoid confounding due to malignancy-related renal dysfunction or treatment effects (e.g., chemotherapy, cachexia), patients with cancer diagnoses (ICD codes beginning with C) were excluded, yielding a final study population of 1240 unique ICU stays.

\begin{figure}[H]
\centering
\includegraphics[width=0.6\linewidth]{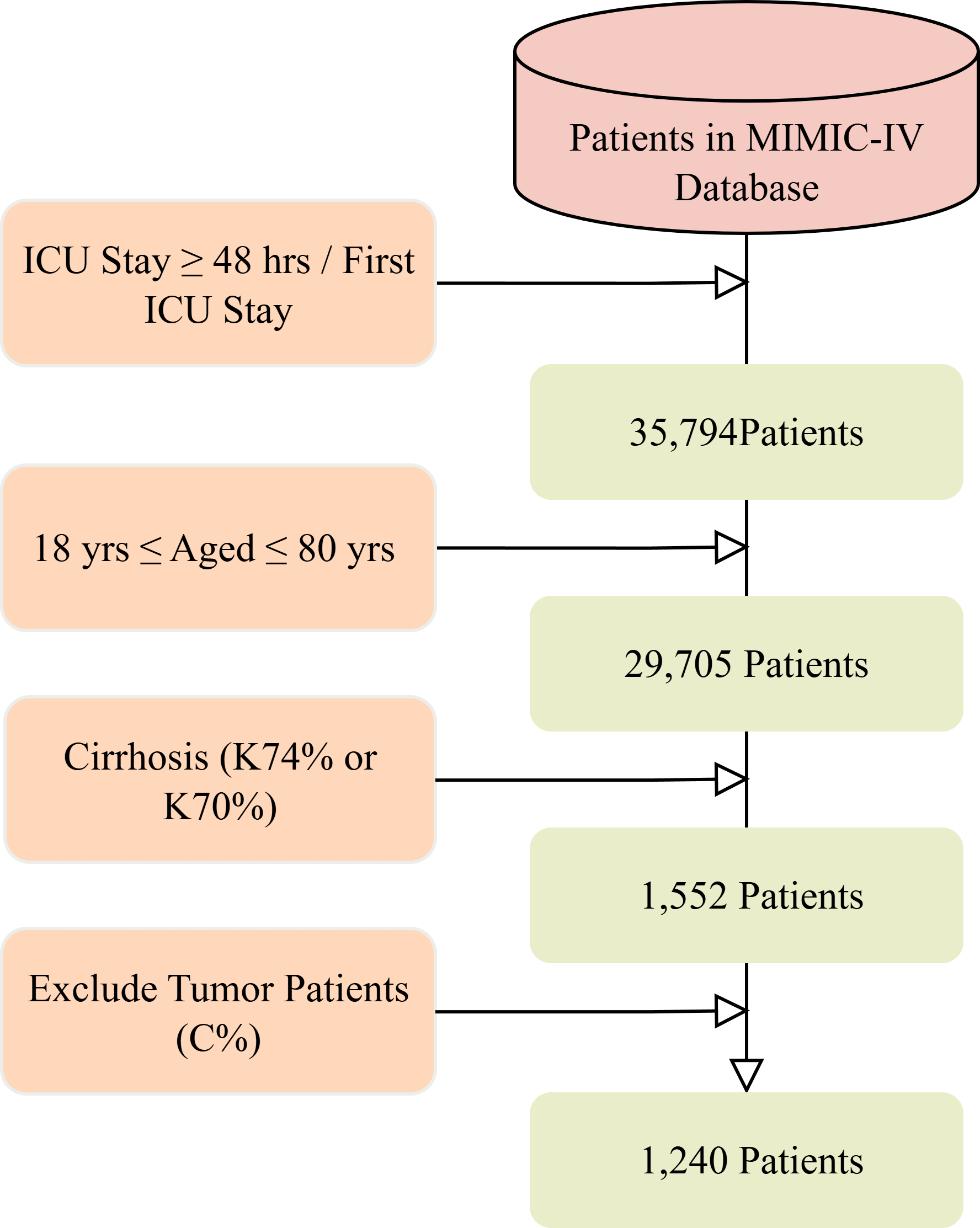}
\caption{Cohort selection process for cirrhosis-associated creatinine elevation analysis}
\label{fig:patient_selection}
\end{figure}

\subsection{Data Preprocessing}

To ensure both statistical robustness and clinical relevance, we applied feature-specific preprocessing strategies based on each variable’s unit, type, and physiologic behavior.

Continuous laboratory variables, including ALT, Hematocrit, Hemoglobin, WBC, were imputed using a k-nearest neighbors  algorithm\cite{abdala2004estimation}. This approach preserves multivariate dependencies and accommodates nonlinear associations commonly observed in liver failure. Each continuous feature was subsequently standardized using \textit{z-score normalization}, which offers robustness to skewed distributions and avoids distortion in models sensitive to feature variance.

To capture early physiologic status while reducing noise from transient fluctuations, we computed the average value within the first 48 hours of ICU admission for selected features\cite{sun2025clinically}. For instance, the \textit{mean pH} and \textit{mean total bilirubin} were used to reflect overall acid-base balance and hepatic excretory function during the early critical phase.

Categorical variables, such as the binary indicator \textit{“20 Gauge placed in outside facility”}, were encoded as 0 or 1. Missing values in such features were treated as absence (0) to reduce false signal amplification caused by imputation bias\cite{sun2025optimizing}.

Given the moderate class imbalance between creatinine elevation and non-elevation groups, we employed the \textit{Synthetic Minority Oversampling Technique} within the training folds\cite{fan2025prediction}. SMOTE synthetically generates minority-class instances by interpolating between existing samples, helping the model better learn decision boundaries while minimizing overfitting. Specifically, for a minority instance $x_{\text{minority}}$ and its $k$-nearest neighbor $x_{\text{neighbor}}$, a synthetic sample is computed as:
\begin{equation}
x_{\text{new}} = x_{\text{minority}} + \delta \cdot (x_{\text{neighbor}} - x_{\text{minority}}), \quad \delta \sim U(0,1)
\end{equation}

All transformations, including imputation, normalization, and SMOTE oversampling, were confined to the training data and subsequently applied to the validation sets. This design ensures proper temporal alignment, eliminates information leakage, and guarantees that model evaluation reflects true generalization performance on unseen ICU patients.

\subsection{Feature Selection}

We employed a two-stage feature selection approach that combines missingness filtering with penalized regression–based sparsity. This strategy was designed to reduce dimensionality, enhance model interpretability, and prevent overfitting in the context of physiologically complex ICU populations. The process began with over 200 candidate features extracted from demographic records, laboratory measurements, clinical interventions, and comorbidity indices. These features were derived from the first 48 hours of ICU admission and initially reviewed by board-certified hepatologists and intensivists to ensure domain relevance and data validity\cite{guo2024association,lin2024predictive}.

\textbf{Stage 1: Missingness Filtering}

We excluded all variables with more than 20\% missingness across the cohort to ensure data quality and minimize bias introduced through imputation. This filtering step removed several high-value but sparsely recorded variables, such as AST, INR, and specific respiratory support parameters. The retained set of features demonstrated sufficient coverage across the population and preserved core indicators of hepatic, renal, and systemic function.

\textbf{Stage 2: Penalized Regression-Based Feature Shrinkage (Lasso)}

To perform multivariate selection and induce sparsity, we applied an L1-regularized logistic regression model to the filtered features. Lasso imposes a penalty on the absolute size of regression coefficients, shrinking the weights of less informative predictors to zero and thereby performing variable selection within the model training process\cite{huang2022lasso}:

\begin{equation}
\hat{\beta} = \arg\min_{\beta} \left\{ \mathcal{L}(\beta) + \lambda \sum_{j=1}^{p} |\beta_j| \right\}
\end{equation}

where \( \mathcal{L}(\beta) \) is the negative log-likelihood of the logistic loss, \( \beta_j \) denotes the coefficient for predictor \( j \), and \( \lambda \) controls the strength of the penalty. We optimized \( \lambda \) through 5-fold cross-validation to identify the minimal subset of variables that preserved predictive performance.

Several features that were retained after missingness filtering but subsequently eliminated by Lasso regularization due to limited marginal contribution included: \textit{Braden Friction/Shear Score}, \textit{Base Excess}, and \textit{Gender}. Although these variables have established clinical relevance in broader ICU populations, they demonstrated minimal additive predictive value in the presence of stronger, cirrhosis-specific indicators and were thus excluded from the final model.

This  process selected 16 final features spanning hepatic biomarkers, systemic indicators, inflammatory markers, and patient-level characteristics. These features are listed in Table~\ref{tab:final_features_cirrhosis} and collectively provide a concise, interpretable foundation for modeling renal injury risk in critically ill cirrhotic patients.

\begin{table}[H]
\centering
\caption{Final selected features used for predicting cirrhosis-associated creatinine elevation}
\label{tab:final_features_cirrhosis}
\renewcommand{\arraystretch}{1.2}
\small
\begin{tabular}{p{4cm}|p{8cm}}
\hline
\rowcolor[HTML]{f7e1d7}
\textbf{Category} & \textbf{Selected Features} \\
\hline
Chartevents & 
ALT, Hematocrit, Hemoglobin, WBC, Anion gap, Admission Weight, Albumin \\
\hline
Labevents & 
PTT, pO\textsubscript{2}, pH, Bilirubin (Total), Calcium (Total), I \\
\hline
Demographic & 
Age, Charlson Comorbidity Index \\
\hline
Procedure/Intervention & 
20 Gauge placed in outside facility \\
\hline
\end{tabular}
\end{table}

These features were further reviewed and validated by clinical experts to ensure medical interpretability, relevance, and non-redundancy.

Each retained feature carries distinct clinical implications in the context of cirrhosis-associated renal risk. \textit{Age} and \textit{Charlson Comorbidity Index} establish baseline frailty and chronic disease burden, which modulate renal vulnerability in liver failure. \textit{Admission Weight} provides an indirect indicator of nutritional status and volume reserve. Liver-specific laboratory markers such as \textit{ALT}, \textit{Albumin}, and \textit{Total Bilirubin} reflect hepatic inflammation, synthetic dysfunction, and excretory impairment, respectively. Electrolyte and acid-base indicators—\textit{Anion Gap}, \textit{pH}, \textit{Calcium}, and \textit{pO\textsubscript{2}}—capture systemic derangement commonly seen in acute decompensation. \textit{WBC}, \textit{PTT}, and \textit{Hematocrit} reflect inflammatory activity, coagulation abnormalities, and hemodilution, all of which are relevant in hepatorenal syndromes. \textit{Iodine}, though less commonly interpreted directly, likely represents metabolic or contrast-related renal influences. Lastly, the presence of a \textit{20 Gauge catheter placed in an outside facility} serves as a surrogate for pre-ICU procedural exposure, reflecting preexisting severity and possibly contributing to fluid shifts or early nephrotoxin exposure.

Together, these 16 features form a compact, interpretable, and clinically validated input space that supports accurate and generalizable modeling of renal risk in cirrhotic ICU populations in Table \ref{tab:feature_definitions}.

\begin{table}[H]
\caption{Detailed Feature Definitions and Encoding Methods}
\label{tab:feature_definitions}
\small
\renewcommand{\arraystretch}{1.2}
\begin{tabularx}{\textwidth}{p{2.5cm}|X|p{1.5cm}|X}
\hline
\rowcolor[HTML]{f7e1d7}
\textbf{Feature} & \textbf{Definition} & \textbf{Units} & \textbf{Source/Calculation} \\
\hline 
AGE & Patient age at ICU admission & years & Calculated from date of birth \\
\hline
Charlson Comorbidity Index & Index summarizing comorbidity burden & score & ICD-based score on admission \\
\hline
ALT & Alanine aminotransferase level & U/L & Charted lab result within first 24h \\
\hline
Hematocrit & Percentage of blood volume occupied by red cells & \% & Charted lab result within first 24h \\
\hline
Hemoglobin & Hemoglobin concentration in blood & g/dL & Charted lab result within first 24h \\
\hline
WBC & White blood cell count & $\times 10^3$/µL & Charted lab result within first 24h \\
\hline
Anion gap & Difference between measured and calculated cations/anions & mmol/L & Charted lab result within first 24h \\
\hline
Admission Weight & Patient weight at ICU admission & Kg & Admission assessment \\
\hline
Albumin & Serum albumin concentration & g/dL & Charted lab result within first 24h \\
\hline
PTT & Partial thromboplastin time & sec & Lab test from labevents \\
\hline
pO\textsubscript{2} & Arterial partial pressure of oxygen & mmHg & Lab test from labevents \\
\hline
pH & Blood acidity/alkalinity level & -- & Lab test from labevents \\
\hline
Bilirubin, Total & Total serum bilirubin concentration & mg/dL & Lab test from labevents \\
\hline
Calcium, Total & Total serum calcium concentration & mg/dL & Lab test from labevents \\
\hline
I & Serum iodine level & µg/dL & Lab test from labevents \\
\hline
20 Gauge placed in outside facility & Whether patient had a 20G catheter placed before hospital transfer & binary & Charted binary intervention field \\
\hline
\end{tabularx}
\end{table}

\subsection{Modeling}

To explore diverse modeling paradigms for predicting creatinine elevation risk in cirrhotic ICU patients, we developed a machine learning framework that incorporates six representative classifiers. These models were chosen to span a wide spectrum of algorithmic complexity, interpretability, and inductive bias—ensuring a comprehensive comparison across both linear and non-linear predictive architectures.

Three gradient-boosted ensemble models— CatBoost, LightGBM, and XGBoost—were included due to their proven effectiveness on structured clinical data\cite{meng2023aspirin}. These methods capture non-linear relationships, handle missing values natively, and account for feature interactions that are often present in critical care physiology. In particular, CatBoost was selected for its ordered boosting scheme and resistance to overfitting in small-to-moderate datasets. LightGBM offers computational efficiency and scalability through leaf-wise tree growth and histogram binning. XGBoost was favored for its flexible regularization and fine-grained control over tree construction, making it suitable for balancing bias–variance tradeoffs.

To anchor model development in interpretability, we implemented a Logistic Regression classifier with both L1 (lasso) and L2 (ridge) regularization\cite{wei2023machine}. As a generalized linear model, logistic regression enables direct coefficient interpretation, making it ideal for clinical deployment scenarios where transparency is essential. It also provides a performance baseline against which more complex models can be evaluated.

A Gaussian Naïve Bayes classifier was introduced to serve as a lightweight probabilistic benchmark\cite{miasnikof2015naive}. Its strong independence assumptions and simplicity make it appealing in low-resource or time-sensitive settings. While often underpowered compared to ensemble methods, its parametric formulation and rapid training allow for effective baseline evaluation.

To assess high-capacity, non-linear representations, we also included a shallow Neural Network with a single hidden layer\cite{gabitashvili2025predicting}. While less interpretable than other models, neural networks are capable of modeling complex interactions and latent representations that may not be captured by tree-based or linear classifiers. The architecture was intentionally kept shallow to prevent overfitting given the dataset size and feature dimensionality.

All models were trained using five-fold stratified cross-validation to ensure robustness and generalizability while avoiding information leakage. Stratified sampling preserved outcome balance across folds, which is particularly important in the setting of class imbalance.

\textbf{Model Evaluation:} We primarily evaluated model discrimination using the area under the receiver operating characteristic curve, a threshold-independent metric well-suited for imbalanced binary classification tasks. AUROC provides a comprehensive assessment of a model’s ability to distinguish between patients at high vs. low risk of renal injury, regardless of any specific clinical cutoff. To account for uncertainty in estimation, 95\% confidence intervals were computed via bootstrapping.

In addition to AUROC, we reported clinically interpretable metrics including accuracy, sensitivity, specificity, F1-score, positive predictive value, and negative predictive value. These metrics collectively address various aspects of clinical decision-making: sensitivity and NPV reflect the ability to avoid missed high-risk cases, while specificity and PPV relate to minimizing overtreatment. The F1-score provides a balanced measure of precision and recall, particularly relevant in the context of limited resources or selective follow-up testing.

In the context of cirrhosis-related renal risk, this multidimensional evaluation strategy is clinically essential. Patients with liver dysfunction often exhibit atypical physiology, making conventional thresholds for renal injury less reliable. A high sensitivity ensures that subtle but clinically meaningful renal deterioration is not overlooked—especially critical in cirrhotic populations where delayed intervention can rapidly lead to hepatorenal syndrome or multiorgan failure. 

Meanwhile, strong NPV performance supports safe clinical de-escalation in patients deemed low risk, potentially reducing unnecessary monitoring or nephrology consultation. On the other hand, specificity and PPV are vital to avoid over-alerting clinicians in resource-constrained ICU settings, where overtreatment can compete with attention to other critically ill patients. 

By incorporating these complementary metrics, our evaluation framework ensures that each model’s clinical utility is interpreted not only through statistical discrimination but also through operational relevance in complex ICU environments.

\subsection{Statistical Analyses}

To ensure the methodological integrity, interpretability, and clinical relevance of our predictive framework for cirrhosis-associated renal risk, we implemented a five-part statistical analysis strategy. This encompassed cohort comparability assessment, feature-level attribution, nonlinear pattern discovery, model explainability, and performance evaluation—each paired with clinical considerations for real-world deployment.

\textbf{1. Cohort Comparability Testing} \\
We tested the statistical equivalence between training and test cohorts using independent two-sided t-tests on continuous features. The test statistic is defined as:

\begin{equation}
t = \frac{\bar{x}_1 - \bar{x}_2}{\sqrt{\frac{s_1^2}{n_1} + \frac{s_2^2}{n_2}}}
\end{equation}

where \( \bar{x}_i \), \( s_i^2 \), and \( n_i \) denote the mean, variance, and sample size in group \( i \). Welch’s correction was applied for unequal variances. This statistical check validates that outcome differences are model-driven rather than artifacts of uneven data splitting. By confirming baseline comparability, we ensure that the model learns generalizable physiologic patterns rather than cohort-specific biases—an essential prerequisite for reliable clinical application across ICUs.

\textbf{2. Marginal Feature Contribution via Ablation} \\
To quantify the standalone predictive power of each variable, we conducted ablation by iteratively removing a feature \( x_i \) and retraining the model\cite{weng2025comparison}. The marginal contribution was computed as the AUROC change:

\begin{equation}
\Delta AUC(x_i) = AUC_{\text{full}} - AUC_{-x_i}
\end{equation}

This allowed identification of features with the greatest impact on prediction, supporting dimensionality reduction and interpretability. Highlighting variables with strong marginal effects (e.g., bilirubin or pH) helps prioritize high-yield clinical markers for nephrotoxicity screening, especially when lab access is limited or rapid risk stratification is needed.

\textbf{3. Accumulated Local Effects} \\
We used ALE plots to visualize nonlinear effects of continuous features while accounting for feature interdependence\cite{chen2025machine}. For feature \( x_j \), ALE at point \( z \) is:

\begin{equation}
ALE_j(z) = \int_{z_0}^{z} \mathbb{E}_{X_{\setminus j}} \left[ \frac{\partial f(X)}{\partial x_j} \Big| x_j = s \right] ds
\end{equation}

Unlike partial dependence plots, ALE avoids extrapolation and handles multicollinearity, which is common in ICU datasets. ALE plots help clinicians understand physiological thresholds where renal injury risk sharply escalates (e.g., bilirubin beyond 5 mg/dL), guiding proactive interventions such as early fluid resuscitation or dose adjustments.

\textbf{4. SHAP-Based Interpretability.} \\
SHAP values decompose a model prediction into additive contributions of each feature\cite{zheng2025explainable}. The SHAP value for feature \( x_i \) is:

\begin{equation}
\phi_i = \sum_{S \subseteq F \setminus \{x_i\}} \frac{|S|!(|F|-|S|-1)!}{|F|!} \left[f(S \cup \{x_i\}) - f(S)\right]
\end{equation}

This provides both global importance rankings and individualized explanations for patient-level predictions. SHAP improves bedside interpretability by linking risk predictions to specific clinical inputs—e.g., identifying that a patient's renal risk is largely driven by hypoalbuminemia—thus supporting explainable and defensible decision-making.

\textbf{5. Multi-Metric Model Evaluation.} \\
We evaluated performance using several complementary metrics:

\begin{itemize}
  \item AUROC: threshold-independent discrimination
  \item Accuracy: overall correctness of classification
  \item Sensitivity: \( \frac{\text{TP}}{\text{TP} + \text{FN}} \), ability to detect renal risk
  \item Specificity: \( \frac{\text{TN}}{\text{TN} + \text{FP}} \), ability to avoid false alarms
  \item F1-score: harmonic mean of precision and recall
  \item PPV: likelihood that a positive prediction is correct
  \item NPV: likelihood that a negative prediction is correct
\end{itemize}

These metrics support balanced model assessment in scenarios where both missed diagnoses (low sensitivity) and over-alerting (low specificity) have serious consequences. NPV and sensitivity are vital for safe discharge or de-escalation, while PPV and specificity reduce overtreatment in already complex cirrhotic ICU cases.

Together, these analytical components form a cohesive evaluation framework that integrates statistical rigor with clinical practicality. By validating dataset integrity, isolating predictive drivers, visualizing nonlinear patterns, enabling model explainability, and balancing evaluation metrics across use-case priorities, our approach ensures that the model is not only technically sound but also clinically interpretable and operationally deployable. This comprehensive strategy lays the foundation for reliable, transparent, and actionable risk prediction of renal injury in cirrhotic ICU patients.

\section{Results}
\subsection{Cohort Characteristics and Statistical Comparison}

The study cohort consists of critically ill cirrhosis patients who were admitted to the ICU, representing a high-risk population with complex hepatorenal interactions. This study specifically focuses on identifying patient profiles that are more prone to developing significant creatinine elevation during their ICU stay, with particular attention to the unique pathophysiological mechanisms underlying kidney dysfunction in advanced liver disease. Although the creatinine changes observed meet the diagnostic thresholds typically used for AKI, the research target is the creatinine response associated with cirrhosis-related complications, not the broader clinical syndrome of AKI. This distinction emphasizes the interest in hepatorenal syndrome, prerenal azotemia from volume depletion, and direct nephrotoxic effects of liver dysfunction rather than all-cause AKI.

Cirrhosis patients in the ICU setting face multiple interconnected challenges that predispose them to renal dysfunction. Portal hypertension leads to splanchnic vasodilation and effective arterial blood volume depletion, triggering compensatory mechanisms including activation of the renin-angiotensin-aldosterone system. These neurohumoral responses initially maintain systemic blood pressure but progressively compromise renal perfusion as liver disease advances. Additionally, the frequent use of diuretics for ascites management, paracentesis procedures, and potential exposure to nephrotoxic medications further compounds the renal vulnerability in this population.

The dataset was randomly divided using stratified sampling into a training set (70\%) and a test set (30\%), ensuring that the distribution of creatinine elevation events was balanced across both subsets. This partitioning strategy is particularly crucial in cirrhosis patients given the heterogeneity of disease severity, ranging from compensated cirrhosis with minimal physiological derangement to decompensated disease with multiorgan dysfunction. The stratified approach minimizes sampling bias related to Child-Pugh class distribution and enhances the reliability and generalizability of the model across different stages of liver disease progression.

As shown in Table~\ref{tab:cirrhosis_cohort_comparison}, none of the 16 clinical variables demonstrated statistically significant differences ($p > 0.05$) between the training and test sets, confirming the internal consistency and comparability of the two cohorts. This statistical equivalence is particularly important given the wide spectrum of laboratory abnormalities typically observed in cirrhosis patients, including coagulopathy (reflected in prolonged PTT), electrolyte imbalances, and altered protein synthesis markers.

\begin{table}[H]
\caption{T-test Comparison of Feature Distributions between Training and Test Sets.}
\label{tab:cirrhosis_cohort_comparison}
\small
\renewcommand{\arraystretch}{1.2}
\rowcolors{2}{white}{white}
\begin{tabularx}{\textwidth}{l|l|X|X|l}
\hline
\rowcolor[HTML]{f7e1d7}
\textbf{Feature} & \textbf{Unit} & \textbf{Training Set} & \textbf{Test Set} & \textbf{P-value} \\ \hline
pH & -- & 7.11 (0.36) & 7.09 (0.32) & 0.314 \\ \hline
PTT & sec & 43.15 (16.42) & 42.83 (16.58) & 0.757 \\ \hline
pO2 & mmHg & 100.25 (50.18) & 103.86 (51.69) & 0.255 \\ \hline
ALT & U/L & 174.61 (563.87) & 151.86 (414.09) & 0.429 \\ \hline
I & µg/dL & 9.39 (11.36) & 8.83 (10.77) & 0.406 \\ \hline
Calcium, Total & mg/dL & 8.33 (0.78) & 8.33 (0.74) & 0.876 \\ \hline
Bilirubin, Total & mg/dL & 7.84 (9.23) & 7.35 (8.83) & 0.376 \\ \hline
WBC & $\times$10\textsuperscript{3}/µL & 12.95 (8.19) & 12.95 (8.37) & 0.992 \\ \hline
Admission Weight & Kg & 86.40 (25.23) & 87.10 (26.85) & 0.669 \\ \hline
Anion Gap & mmol/L & 15.71 (5.11) & 15.73 (5.17) & 0.950 \\ \hline
Albumin & g/dL & 2.98 (0.54) & 3.01 (0.48) & 0.263 \\ \hline
Hematocrit & \% & 27.73 (5.76) & 28.21 (6.04) & 0.194 \\ \hline
Hemoglobin & g/dL & 9.15 (1.88) & 9.34 (2.01) & 0.126 \\ \hline
20 Gauge Outside Facility & binary & 0.29 (0.37) & 0.25 (0.33) & 0.063 \\ \hline
Age & years & 55.71 (12.78) & 56.67 (12.18) & 0.210 \\ \hline
Charlson Comorbidity Index & score & 5.27 (2.50) & 5.26 (2.32) & 0.927 \\ \hline
\end{tabularx}
\begin{flushleft}
\textit{Note}: This table summarizes statistical comparisons between the training and test cohorts. Continuous variables are expressed as mean (standard deviation). P-values are derived from two-sided t-tests, with significance set at $p<0.05$.
\end{flushleft}
\end{table}

Table~\ref{tab:cirrhosis_elevation_comparison} presents the baseline characteristics of the creatinine elevation and non-elevation groups, revealing distinct pathophysiological patterns. Patients with creatinine elevation exhibited significantly higher levels of PTT (47.46 vs 41.65 seconds, p<0.001), reflecting more severe coagulopathy and advanced hepatic synthetic dysfunction. Higher ALT levels (268.10 vs 142.09 U/L, p=0.016) and markedly elevated total bilirubin (10.94 vs 6.76 mg/dL, p<0.001) in the creatinine elevation group indicate more severe hepatocellular injury and cholestasis, suggesting advanced liver dysfunction that predisposes to hepatorenal syndrome through complex neurohumoral mechanisms.

Patients with creatinine elevation showed lower pH values (7.04 vs 7.14, p<0.001), indicating metabolic acidosis that may result from impaired lactate clearance by the diseased liver, reduced renal acid excretion, or tissue hypoperfusion. The significantly lower hematocrit (26.18\% vs 28.27\%, p<0.001) and hemoglobin levels (8.69 vs 9.31 g/dL, p<0.001) reflect the multifactorial anemia common in cirrhosis, including gastrointestinal bleeding, hypersplenism from portal hypertension, and chronic disease-related bone marrow suppression.

These findings are consistent with known mechanisms of cirrhosis-associated nephrotoxicity, including hepatorenal syndrome type 1 (rapid deterioration) and type 2 (gradual decline), prerenal azotemia from effective volume depletion, and acute tubular necrosis from nephrotoxic exposures or hemodynamic instability. By focusing specifically on creatinine elevation linked to cirrhosis complications, this study offers a targeted perspective on hepatic-renal interactions that extends beyond traditional assessments of liver function alone. The internal consistency of the dataset, along with the biological plausibility of the identified predictors reflecting well-established pathophysiological mechanisms in advanced liver disease, supports the clinical relevance and reliability of the proposed model in assessing cirrhosis-associated creatinine elevation risk in the critical care setting.

\begin{table}[H]
\caption{T-test Comparison of Feature Distributions Between Non-Elevation and Elevation Groups.}
\label{tab:cirrhosis_elevation_comparison}
\small
\renewcommand{\arraystretch}{1.2}
\rowcolors{2}{white}{white}
\begin{tabularx}{\textwidth}{l|l|X|X|l}
\hline
\rowcolor[HTML]{f7e1d7}
\textbf{Feature} & \textbf{Unit} & \textbf{Non-Elevation Set} & \textbf{Elevation Set} & \textbf{P-value} \\ \hline
pH & -- & 7.14 (0.36) & 7.04 (0.33) & < 0.001 \\ \hline
PTT & sec & 41.65 (16.17) & 47.46 (16.38) & < 0.001 \\ \hline
pO2 & mmHg & 96.11 (49.67) & 112.15 (49.84) & < 0.001 \\ \hline
ALT & U/L & 142.09 (494.48) & 268.10 (720.62) & 0.016 \\ \hline
I & µg/dL & 8.08 (10.48) & 13.17 (12.86) & < 0.001 \\ \hline
Calcium, Total & mg/dL & 8.25 (0.78) & 8.55 (0.73) & < 0.001 \\ \hline
Bilirubin, Total & mg/dL & 6.76 (8.35) & 10.94 (10.84) & < 0.001 \\ \hline
WBC & $\times$10\textsuperscript{3}/µL & 12.30 (7.75) & 14.80 (9.11) & < 0.001 \\ \hline
Admission Weight & Kg & 85.17 (24.59) & 89.93 (26.74) & 0.020 \\ \hline
Anion Gap & mmol/L & 15.43 (5.23) & 16.51 (4.65) & 0.004 \\ \hline
Albumin & g/dL & 2.95 (0.52) & 3.05 (0.60) & 0.031 \\ \hline
Hematocrit & \% & 28.27 (5.99) & 26.18 (4.71) & < 0.001 \\ \hline
Hemoglobin & g/dL & 9.31 (1.96) & 8.69 (1.53) & < 0.001 \\ \hline
20 Gauge Outside Facility & binary & 0.34 (0.39) & 0.17 (0.26) & < 0.001 \\ \hline
Age & years & 56.31 (12.73) & 54.01 (12.83) & 0.021 \\ \hline
Charlson Comorbidity Index & score & 5.27 (2.51) & 5.27 (2.48) & 0.990 \\ \hline
\end{tabularx}
\begin{flushleft}
\textit{Note}: This table compares patients with and without cirrhosis-associated creatinine elevation. Continuous features are presented as mean (standard deviation). P-values were calculated using two-sided t-tests with a significance threshold of $p<0.05$. The creatinine elevation events included in this study meet established diagnostic thresholds commonly used for AKI but specifically represent cirrhosis-associated renal response.
\end{flushleft}
\end{table}

\subsection{Feature Contribution Analysis}

\begin{figure}[H]
    \centering
    \includegraphics[width=1\linewidth]{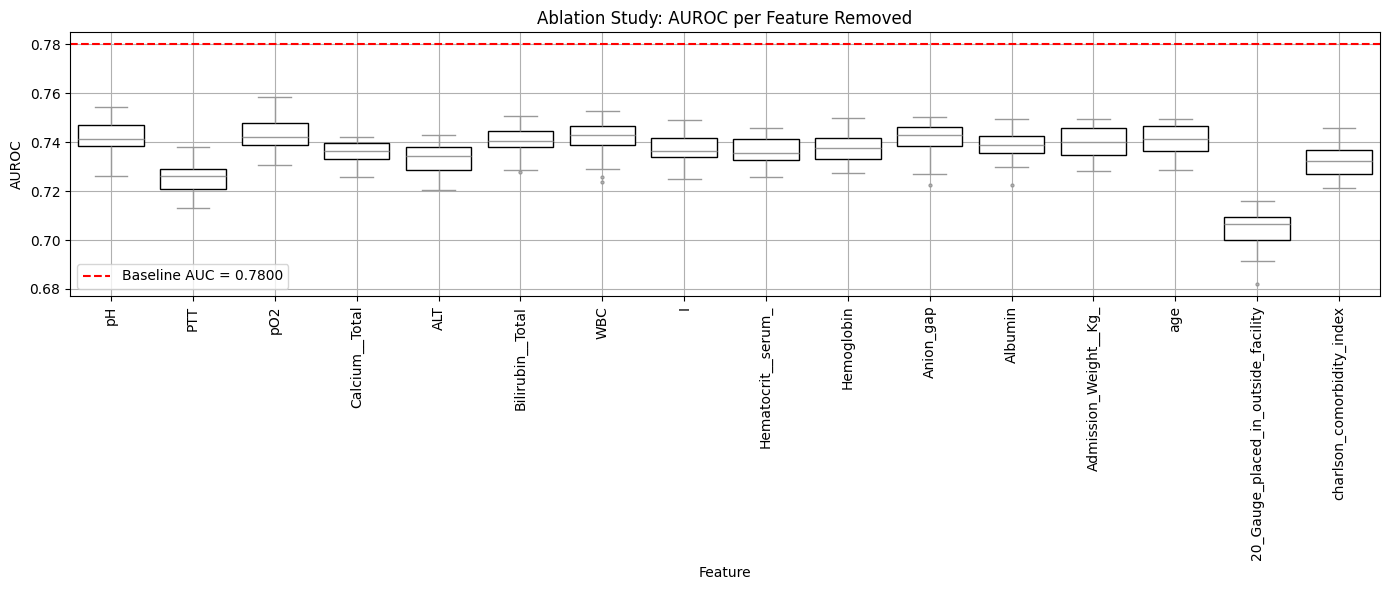}
    \caption{Impact of Feature Removal on LR Model Performance for Cirrhosis Patients.}
    \label{fig:cirrhosis_ablation}
\end{figure}

To evaluate individual characteristic contributions to predict creatinine elevation associated with cirrhosis in critically ill patients, we performed an ablation analysis. As illustrated in Figure~\ref{fig:cirrhosis_ablation}, each feature was systematically removed, and a logistic regression classifier was retrained using bootstrap sampling to assess its marginal impact on AUROC. The red dashed line represents the baseline AUROC (0.780) achieved when all features were included. This analysis aimed to quantify the predictive utility of each variable independently of specific model architectures, providing information on which clinical factors most strongly contribute to the evaluation of nephrotoxicity risk.

Notably, the exclusion of 20 gauge placement in outside facility and PTT led to the most substantial decline in model performance, indicating that these factors play dominant roles in identifying patients at higher risk of creatinine elevation in cirrhosis patients. The 20 gauge placement reflects procedural urgency and disease acuity, as patients requiring immediate large-bore vascular access often present with acute decompensation events such as variceal bleeding, which inherently predispose to renal hypoperfusion. This is consistent with clinical observations that procedural factors and coagulation abnormalities may reflect disease severity and hemodynamic instability, which are critical in the development of hepatorenal complications.

The dominance of PTT as a predictive feature underscores the central role of hepatic synthetic function in determining renal risk. Prolonged PTT reflects impaired synthesis of vitamin K-dependent clotting factors and indicates advanced hepatocellular dysfunction that correlates with the Child-Pugh class and the severity of the MELD score.

Other features with considerable impact included pH and pO2, which showed significant performance decreases when removed, highlighting the importance of acid-base status and oxygenation in the assessment of nephrotoxicity risk. The predictive significance of pH reflects early metabolic acidosis from impaired hepatic lactate clearance and reduced renal acid excretion, serving as an integrated marker of both hepatic and early renal dysfunction. In contrast, features such as charlson comorbidity index and several other laboratory values showed minimal impact when excluded, suggesting these variables provide limited unique predictive information in this clinical context.

\subsection{Model Performance on Creatinine Elevation Risk Prediction}

\begin{figure}[H]
\centering
\includegraphics[width=0.75\linewidth]{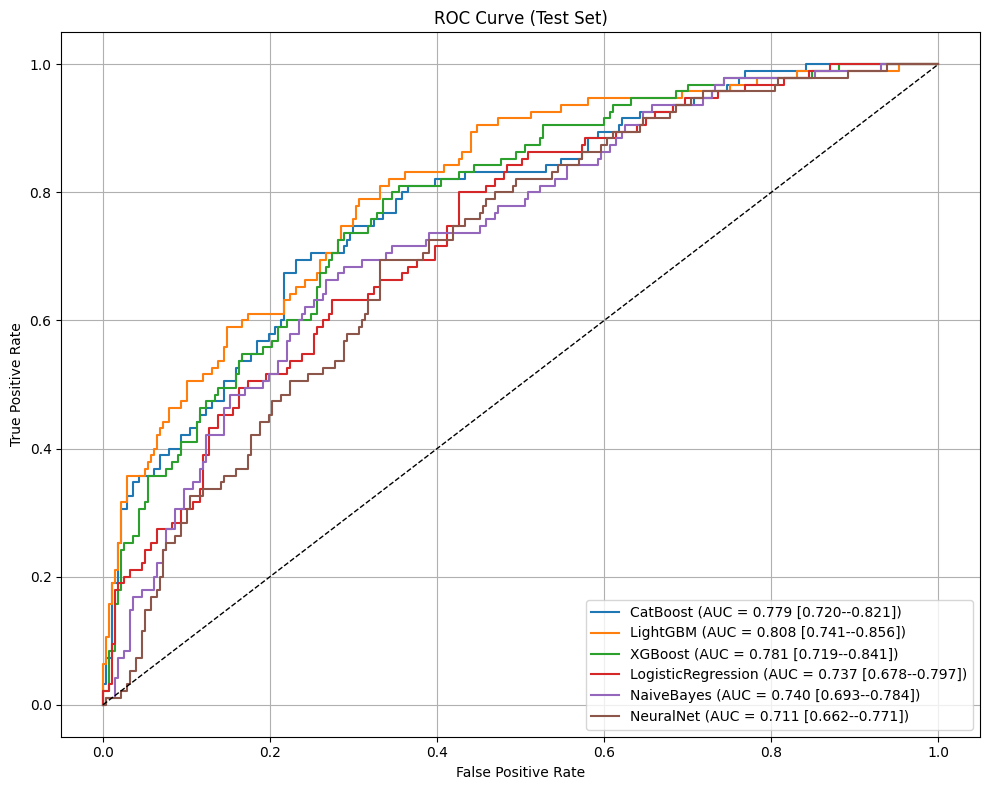}
\caption{AUROC Curves for Model Performance in Cirrhosis Patients Test Set.}
\label{fig:cirrhosis_roc}
\end{figure}

To evaluate the capacity of different algorithms to predict cirrhosis-associated creatinine elevation among ICU patients, we tested six widely used machine learning models across a comprehensive range of performance metrics. Performance metrics on the test set—including AUROC, sensitivity, specificity, F1-score, and predictive values—are summarized in Table~\ref{tab:cirrhosis_results}. ROC curves for the test set are illustrated in Figure~\ref{fig:cirrhosis_roc}.

\begin{table}[H]
\renewcommand{\arraystretch}{1.2}
\centering
\caption{Performance Comparison of Different Models for Cirrhosis Patients in the Test Set.}
\resizebox{\textwidth}{!}{
\begin{tabular}{l|l|l|l|l|l|l|l}
\hline
\rowcolor[HTML]{f7e1d7}
\textbf{Model} & \textbf{AUROC (95\% CI)} & \textbf{Accuracy} & \textbf{F1-score} & \textbf{Sensitivity} & \textbf{Specificity} & \textbf{PPV} & \textbf{NPV} \\ \hline
\rowcolor[HTML]{a8dadc}
LightGBM & 0.808 (0.741--0.856) & 0.704 & 0.583 & 0.811 & 0.668 & 0.456 & 0.911 \\
XGBoost & 0.781 (0.719--0.841) & 0.691 & 0.569 & 0.800 & 0.653 & 0.442 & 0.905 \\
CatBoost & 0.779 (0.720--0.821) & 0.683 & 0.563 & 0.800 & 0.643 & 0.434 & 0.904 \\
LogisticRegression & 0.737 (0.678--0.797) & 0.632 & 0.526 & 0.800 & 0.574 & 0.392 & 0.893 \\
NaiveBayes & 0.740 (0.693--0.784) & 0.570 & 0.487 & 0.800 & 0.491 & 0.350 & 0.877 \\
NeuralNet & 0.711 (0.662--0.771) & 0.599 & 0.505 & 0.800 & 0.531 & 0.369 & 0.886 \\ \hline
\end{tabular}
}
\label{tab:cirrhosis_results}
\end{table}

Among the models evaluated, LightGBM achieved the highest AUROC of 0.808 (95\% CI: 0.741–0.856), indicating strong discriminatory ability in distinguishing between cirrhosis patients who will and will not develop significant creatinine elevation during their ICU stay. The model's superior performance can be attributed to its efficient gradient boosting framework and native handling of missing data, which is particularly relevant in ICU settings where laboratory values may be incomplete or obtained at irregular intervals. LightGBM also delivered the best overall accuracy (0.704), highest F1-score (0.583), and top negative predictive value (NPV = 0.911), demonstrating robust performance across multiple evaluation dimensions despite the challenging class imbalance.

The exceptional NPV of 0.911 achieved by LightGBM has profound clinical implications in ICU practice. This metric indicates that when the model predicts a cirrhosis patient will not develop creatinine elevation, there is a 91.1\% probability that this prediction is correct. In practical ICU scenarios, this high confidence in ruling out renal risk allows clinicians to focus intensive monitoring resources on truly high-risk patients while maintaining standard care protocols for predicted low-risk cases. This is particularly valuable in resource-constrained ICU environments where continuous renal replacement therapy machines, specialized nursing attention, and nephrology consultations must be allocated efficiently.

CatBoost and XGBoost showed comparable performance with AUROCs of 0.779 and 0.781 respectively, while maintaining solid specificity around 0.64-0.65 at the fixed sensitivity threshold of 0.800. This consistency across gradient boosting methods suggests robust signal detection in the cirrhosis-nephrotoxicity relationship, with each algorithm capturing similar underlying patterns in the data through different optimization strategies. The maintained specificity levels indicate that approximately two-thirds of patients predicted as low-risk will indeed avoid significant creatinine elevation, providing reasonable confidence for clinical decision-making.

The fixed sensitivity threshold of 0.800 was deliberately chosen to reflect clinical priorities in ICU cirrhosis management, where missing a patient destined for renal failure (false negative) carries far greater consequences than triggering enhanced monitoring for a patient who ultimately maintains stable renal function (false positive). This threshold ensures that 80\% of patients who will develop creatinine elevation are correctly identified, allowing for timely interventions such as nephrotoxic medication avoidance, hemodynamic optimization, and early nephrology consultation before irreversible renal damage occurs.

From a clinical perspective, this level of performance is particularly valuable in real-world ICU settings where cirrhosis patients are at high risk for hepatorenal syndrome and other renal complications that can rapidly progress to multiorgan failure. The models' ability to predict creatinine elevation before it becomes clinically apparent provides a crucial window for intervention. In practical ICU scenarios, early identification enables several critical interventions: immediate discontinuation of nephrotoxic medications such as NSAIDs or aminoglycosides, optimization of hemodynamic parameters through careful fluid management and vasoactive support, initiation of renal-protective strategies including avoidance of contrast agents, and proactive nephrology consultation for consideration of early renal replacement therapy if indicated.

The high NPV across all models (>0.87) is particularly useful for clinical decision-making, enabling confident identification of low-risk patients while implementing enhanced monitoring protocols for high-risk patients. This dual-tiered approach optimizes resource allocation in busy ICU environments, where the difference between standard monitoring (routine creatinine checks every 24-48 hours) and intensive monitoring (creatinine checks every 6-12 hours with real-time nephrology involvement) can be the difference between reversible renal dysfunction and irreversible kidney failure requiring long-term dialysis. The model's reliability in ruling out high-risk patients allows ICU teams to confidently apply standard protocols to the majority of cirrhosis patients while concentrating specialized renal protection efforts on the identified high-risk subset.

\subsection{SHAP analysis and feature attribution}

\begin{figure}[H]
    \centering
    \includegraphics[width=0.7\linewidth]{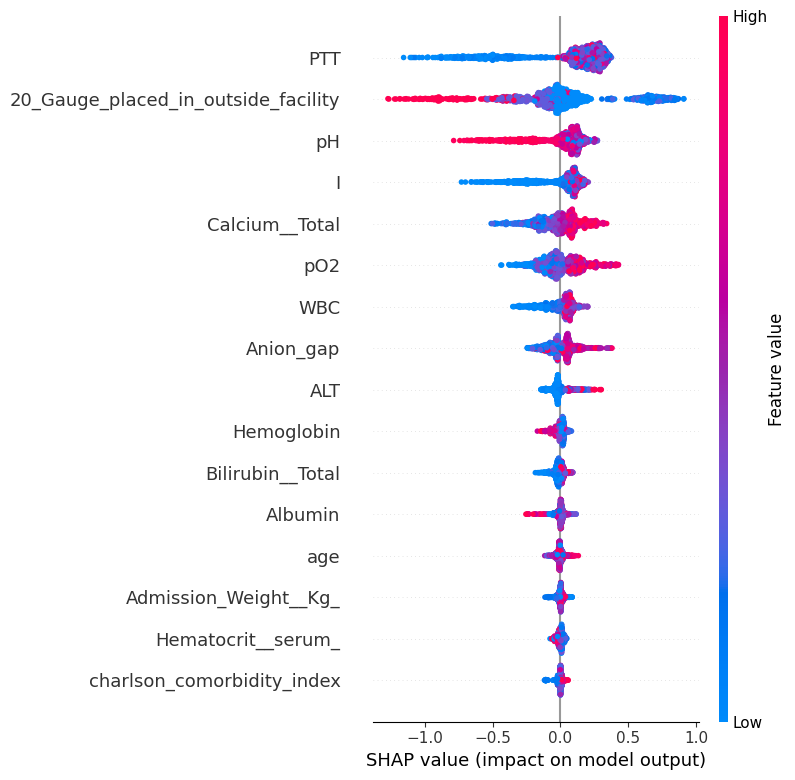}
    \caption{SHAP summary plot showing feature contributions to predicted creatinine elevation in cirrhosis patients.}
    \label{fig:cirrhosis_shap}
\end{figure}

To interpret the contribution of individual variables in predicting cirrhosis-associated creatinine elevation among ICU patients, SHAP analysis was applied to the LightGBM model. Figure~\ref{fig:cirrhosis_shap} presents a SHAP summary plot, where each point represents an individual patient. The x-axis shows the SHAP value, which reflects the degree to which a feature influences the model's prediction for that patient. Colors indicate the feature value: red points correspond to high values, and blue points represent low values.

PTT was the most influential predictor, with high PTT levels (red points) consistently positioned on the right side of the plot, indicating that elevated PTT levels substantially increase the predicted risk of creatinine elevation. This pattern is clinically plausible, as prolonged PTT may signal severe coagulopathy and advanced liver dysfunction in cirrhosis patients. The strong directional relationship suggests that as hepatic synthetic function deteriorates, reflected by impaired production of vitamin K-dependent clotting factors, the risk for hepatorenal syndrome increases proportionally.

The 20 gauge placement in outside facility demonstrated a complex relationship, with low values (blue points) appearing to increase renal risk, likely reflecting more urgent presentations or acute decompensation events that bypass outside stabilization attempts. pH also ranked among the top contributors, suggesting that procedural factors and acid-base status play important roles in risk prediction. Lower pH values consistently increase predicted renal risk, reflecting metabolic consequences of advanced liver disease including impaired lactate clearance and early renal dysfunction.

The model's reliance on physiologically meaningful predictors supports its clinical interpretability and enhances its potential for integration into real-world ICU risk management for cirrhosis patients. The SHAP analysis reveals that the model has learned clinically relevant patterns rather than spurious correlations, providing confidence in its bedside applicability.

\subsection{ALE analysis and clinical interpretability}

\begin{figure}[H]
    \centering
    \includegraphics[width=1\linewidth]{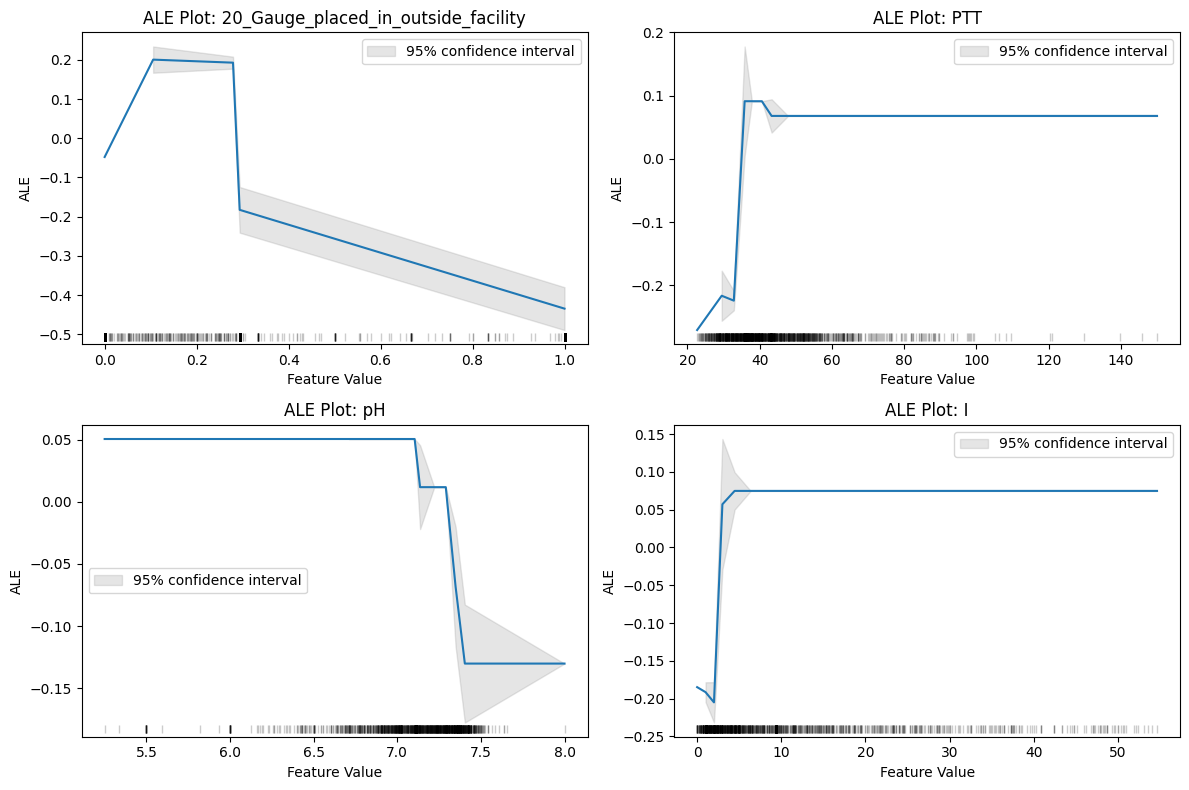}
    \caption{ALE plots for top features in cirrhosis ICU patients.}
    \label{fig:cirrhosis_ale}
\end{figure}

To further investigate the local interpretability and clinical plausibility of our LightGBM model in predicting cirrhosis-associated creatinine elevation among ICU patients, we conducted an ALE analysis focusing on four high-impact features: 20 gauge placement in outside facility, PTT, pH, and I. The ALE plots, shown in Figure~\ref{fig:cirrhosis_ale}, illustrate the marginal effect of each variable on the model's output while accounting for the influence of other features. These visualizations reveal how each clinical parameter influences kidney injury risk across different value ranges in cirrhosis patients.

The 20 gauge placement feature exhibited a complex relationship with risk prediction, showing initial risk increase followed by a plateau and subsequent decrease. This pattern reflects that patients requiring 20-gauge access at outside facilities represent moderate-severity cases needing stabilization before transfer, whereas absence of such placement indicates either stable patients or critically unstable patients transferred emergently without procedural interventions. The highest renal risk occurs in patients who received outside intervention but still required ICU transfer, suggesting partial stabilization of severe decompensation events.

PTT demonstrated a continuous positive relationship with risk, aligning with clinical understanding that prolonged clotting times indicate more severe liver dysfunction. The steep initial rise in renal risk as PTT increases from normal values (around 30 seconds) to mildly prolonged levels (40-50 seconds) reflects the transition from compensated to decompensated liver disease. As PTT extends beyond 60 seconds, the model shows progressively increasing renal risk, corresponding to severe coagulopathy that predisposes to hepatorenal syndrome through bleeding risk and synthetic function failure.

pH showed a relatively stable effect across most physiological ranges with a sharp decrease in renal risk at very high pH values, suggesting that severe alkalosis may represent different underlying pathophysiology. This likely reflects compensatory respiratory alkalosis in patients with hepatic encephalopathy, where hyperventilation maintains acid-base balance. However, the model appropriately identifies that most cirrhosis patients with renal risk present with metabolic acidosis from impaired hepatic lactate clearance or tissue hypoperfusion.

The variable "I" demonstrated a gradual positive relationship with risk prediction, showing that higher values consistently increase creatinine elevation probability. This parameter appears to capture metabolic dysfunction aspects particularly relevant in cirrhosis patients, where ion transport mechanisms may be disrupted by advanced liver disease. The steady increase suggests this marker reflects cumulative metabolic burden rather than a threshold effect.

Overall, the ALE analysis confirms that the model's risk estimations are biologically coherent and sensitive to clinically relevant ranges of key predictors in cirrhosis patients. The identified relationships align with established pathophysiological principles of hepatorenal syndrome, supporting the model's potential for clinical implementation as a decision support tool where early recognition of renal risk can guide timely therapeutic interventions.

\section{Discussion}
\subsection{Summary of Existing Model Compilation}
This study introduces a clinically interpretable and methodologically robust machine learning framework for predicting AKI in critically ill patients with cirrhosis. The modeling pipeline combined structured data preprocessing, a two-stage feature selection strategy (missingness filtering followed by LASSO regularization), and class balancing via SMOTE, with subsequent evaluation across six representative classification algorithms.

Among the tested models, LightGBM achieved the highest discrimination on the test set (AUROC: 0.808; 95\% CI: 0.741–0.856), as well as the highest accuracy (0.704), F1-score (0.583), and negative predictive value (0.911). These findings indicate that the model is well suited to identify low-risk patients with high confidence—an especially valuable capability in ICU settings where early exclusion of AKI risk can help prioritize resources for those most in need. Although logistic regression demonstrated lower predictive accuracy, its inherent transparency and coefficient interpretability make it a valuable reference for clinical decision-making, particularly in environments where model explainability is a critical requirement.

Interpretability was enhanced through ablation analysis, which quantified the marginal impact of each predictor, and SHAP/ALE analyses, which revealed both global and localized feature effects. PTT emerged as the strongest predictor, consistent with prior reports linking prolonged coagulation times to renal dysfunction in cirrhosis \cite{Tian2024}: prolonged PTT reflects impaired hepatic synthetic function and advanced coagulopathy, conditions that are strongly associated with systemic circulatory dysfunction and increased AKI susceptibility in cirrhosis. The absence of outside-facility 20 gauge placement was linked to higher AKI risk, likely reflecting more acute or unstable presentations without prior stabilization. Low pH values indicated metabolic acidosis, often a consequence of impaired lactate clearance and reduced renal acid excretion, both of which signal early renal compromise\cite{Sun2019}. Altered pO$_2$ levels highlighted the interplay between hypoxemia, hemodynamic instability, and renal perfusion deficits\cite{Drolz2018}. These findings emphasize the physiologic plausibility and interventional potential of the predictors identified in our model. The identification of these factors provides not only predictive utility but also direct clinical targets for intervention—such as early correction of coagulopathy, optimization of acid–base status, and maintenance of adequate oxygen delivery.

By leveraging routinely available laboratory and clinical parameters, the proposed model supports real-time AKI risk stratification in an understudied yet high-risk ICU population. The capacity to detect high-risk patients before overt kidney injury develops offers a critical window for nephroprotective measures, including the avoidance of nephrotoxic agents, individualized hemodynamic optimization, and early nephrology involvement. Conversely, confidently identifying low-risk patients allows clinicians to avoid unnecessary invasive monitoring or interventions, thereby reducing patient burden and optimizing ICU resource allocation. This dual benefit—precise targeting of preventive strategies for high-risk patients and safe de-escalation for low-risk patients—positions the model as a clinically actionable decision support tool in the management of cirrhotic patients at risk for AKI.

\subsection{Comparison with Prior Studies}
Multiple studies have examined AKI risk in cirrhosis, yet most either rely on traditional regression with limited nonlinearity capture, focus on general inpatients rather than ICU cohorts, or provide static risk scores that are difficult to operationalize at the bedside. Our work advances this literature along three dimensions: (i) an ICU-focused, early-window feature set tailored to real-time triage; (ii) a modern ML toolkit that models nonlinear interactions while preserving interpretability; and (iii) clinically actionable readouts emphasizing negative predictive value and intervention targets.

Patidar et al.\cite{Patidar2019AKI} derived and externally validated a 3-variable logistic model (admission creatinine, INR, WBC) to predict in-hospital AKI among hospitalized cirrhotics, reporting AUROC 0.77 in derivation and 0.70 in external validation. Unlike their general-ward design, we concentrate on the ICU—where AKI typically emerges within a few days—use richer physiologic inputs from the early ICU phase, and explicitly quantify patient-level explanations. This ICU focus enables more immediate resource prioritization and nephroprotective strategies.

Closer to our setting, Tu et al.\cite{Tu2024} developed a dynamic nomogram for early AKI in ICU cirrhotics using MIMIC data, achieving AUCs of 0.797 (train) and 0.750 (validation). While practical and easy to use, that tool remains a linear scoring framework. Our LightGBM model reached higher discrimination on a held-out test set (AUROC: 0.808) while maintaining bedside interpretability via ablation, SHAP, and ALE. Importantly, we highlight a high NPV, supporting safe de-escalation for low-risk patients and targeted escalation for high-risk ones—two directions that translate directly into ICU workflow.

Several contemporaneous studies address related but distinct questions\cite{Nadim2024}—e.g., prognostic nomograms for mortality in cirrhotics with AKI or remodeling MELD in cirrhosis+AKI—thereby informing downstream outcomes rather than incident AKI risk. Our contribution is complementary: by flagging high-risk cirrhotic patients beforeovert kidney injury, we provide a preventive window for correcting coagulopathy, optimizing acid–base status and oxygen delivery, avoiding nephrotoxins, and involving nephrology early. This aligns with modern consensus guidance emphasizing early recognition and phenotype-aware management of AKI in cirrhosis.

Finally, reviews and consensus statements underscore that AKI in cirrhosis is common, rapidly progressive, and tightly linked to short-term mortality, with HRS-AKI carrying especially high risk. By operationalizing an interpretable, ICU-ready model based on routinely available variables, our study addresses a recognized gap: moving from broad risk awareness to actionable, patient-specific early warnings that can be embedded into critical-care pathways.

\subsection{Clinical Integration and Operationalization}
To translate high model performance into tangible bedside benefits, practical deployment strategies are essential. In a real-world ICU environment, our AKI prediction model could be embedded within the hospital’s electronic health record (EHR) system as part of an automated renal risk dashboard. Predictions would be updated every 6–12 hours using the latest available vital signs and laboratory results—particularly within the first 48 hours of ICU admission, when our results indicate the highest discriminative capacity. The system could stratify patients into high-, intermediate-, and low-risk categories based on predicted probability.

For high-risk patients—identified by factors such as prolonged PTT, metabolic acidosis (low pH), hypoxemia (altered pO$_2$), and absence of prior stabilization (no outside-facility 20G placement)—the dashboard could automatically alert intensivists, hepatologists, and nephrology teams. This would prompt early multidisciplinary huddles to initiate nephroprotective measures, optimize hemodynamics, correct coagulopathy, and avoid nephrotoxic medications. Intermediate-risk patients could undergo intensified monitoring, more frequent laboratory checks, and proactive fluid and medication review. Low-risk patients, given the model’s high negative predictive value (0.911), could be prioritized for standard monitoring protocols or early step-down care, improving ICU bed turnover without compromising safety.

Such an operational framework offers multiple clinical and economic benefits. Early identification of high-risk patients enables timely intervention, potentially reducing progression to severe AKI and associated short-term mortality. Stratified monitoring allows more efficient use of ICU nursing and diagnostic resources, while safe de-escalation for low-risk patients frees capacity for critically ill admissions. For example, if early alerts allow prevention of AKI in even 10\% of flagged high-risk cases, the downstream cost savings could be substantial given the high costs of renal replacement therapy and prolonged ICU stays. Furthermore, by integrating explainable risk drivers into the dashboard, the system may improve clinician trust, facilitate patient–family discussions about prognosis, and support shared decision-making in cases with limited reversibility.

\subsection{Limitations and Future Work}
An important limitation of this study is that the selected feature set—derived from the MIMIC-IV database—differs from those available in most other ICU datasets, preventing direct external validation at this stage. This feature mismatch limits our ability to benchmark the model’s performance across institutions without additional data harmonization efforts. Despite the strengths of our analysis—including the use of a high-resolution ICU dataset, a robust machine learning pipeline, and clinically interpretable feature attribution—several other limitations must be acknowledged. 

First, the analysis is based on retrospective, single-center data from the MIMIC-IV database, which may further limit external generalizability. Although the dataset is large and diverse, it reflects the practices of a specific healthcare system, and clinical management strategies may differ in other regions. External validation on multi-center and prospective cohorts will be essential before clinical deployment.

Second, we restricted model inputs to routinely available laboratory and physiological variables within the first 48 hours of ICU admission. While this supports early AKI risk stratification, it may not capture evolving pathophysiological changes. Incorporating time-series data and trends—such as dynamic coagulation parameters, serial acid–base measurements, and oxygenation indices—could improve the ability to detect rapid deterioration. Additionally, we did not have access to novel renal biomarkers (e.g., NGAL, cystatin C), viscoelastic coagulation testing, or detailed medication exposure data (e.g., diuretics, nephrotoxins), which could further refine predictions.

Third, our feature set, though interpretable, may still be influenced by confounding variables and underlying biases. For example, certain procedural indicators, such as the absence of outside-facility 20G placement, may reflect care delivery patterns rather than purely biological risk. Socio-demographic variables were not included in the present analysis to minimize the risk of perpetuating healthcare disparities; however, equity-focused auditing will be important in future implementation phases.

In future work, we plan to expand in three key directions. From a methodological perspective, we will evaluate temporal deep learning models (e.g., LSTM, transformers) to capture longitudinal deterioration patterns and explore multimodal integration by incorporating imaging (e.g., Doppler ultrasound, elastography) and novel biomarkers. From a clinical perspective, we aim to link ICU data with post-discharge outcomes to predict not only AKI onset but also longer-term endpoints such as renal recovery, dialysis dependence, and survival at 3, 6, and 12 months. Finally, from an implementation standpoint, we will design a user-friendly interface embedded in electronic health record systems, enabling real-time risk scoring, automated alerts, and built-in equity monitoring. Such integration will allow proactive, personalized AKI prevention strategies in critically ill cirrhotic patients while ensuring transparency, fairness, and clinical utility.

\section{Conclusion}
This study developed and validated a clinically interpretable machine learning framework for predicting acute kidney injury in critically ill patients with cirrhosis using routinely available ICU data from the first 48 hours of admission. The LightGBM model demonstrated the highest predictive performance, achieving an AUROC of 0.808 and a negative predictive value of 0.911, enabling both early identification of high-risk patients and safe de-escalation for low-risk patients. Feature contribution analysis revealed physiologically plausible and potentially modifiable predictors—such as prolonged PTT, absence of outside-facility 20G placement, metabolic acidosis, and altered oxygenation—that align with known pathophysiological mechanisms and offer direct interventional targets.

Compared with prior studies, our approach leverages high-resolution ICU data, incorporates dynamic physiologic variables, and applies robust interpretability methods to bridge the gap between statistical accuracy and bedside applicability. The proposed framework can be operationalized within an electronic health record system to provide automated, real-time renal risk stratification, supporting timely nephroprotective interventions, optimized ICU resource allocation, and enhanced shared decision-making.

While these findings highlight the model’s potential clinical value, limitations—including reliance on single-center retrospective data, differences in feature availability across datasets, and the absence of external validation—necessitate further research. Future work should focus on multi-center prospective validation, integration of temporal and multimodal data sources, and the development of equity-aware deployment tools to ensure both performance and fairness in diverse clinical settings.

In summary, this study provides a proof-of-concept for an interpretable, ICU-ready AKI risk prediction tool tailored to cirrhotic patients. By translating high model performance into actionable clinical pathways, it offers a promising step toward improving early detection, prevention, and personalized management of AKI in this vulnerable population.

\section*{Declarations}

\textbf{Funding}  
Not applicable. This research did not receive any specific grant from funding agencies in the public, commercial, or not-for-profit sectors.

\textbf{Conflict of interest/Competing interests}  
The authors declare that they have no known competing financial interests or personal relationships that could have appeared to influence the work reported in this paper.

\textbf{Ethics approval and consent to participate}  
This study used the publicly available MIMIC-IV database \cite{johnson2020mimic}, which has existing Institutional Review Board (IRB) approval from the Massachusetts Institute of Technology and Beth Israel Deaconess Medical Center. All data were de-identified in compliance with the Health Insurance Portability and Accountability Act (HIPAA) standards. Therefore, no additional ethics approval or informed consent was required for this retrospective analysis.

\textbf{Consent for publication}  
Not applicable.

\textbf{Data availability}  
The MIMIC-IV dataset \cite{johnson2020mimic} is publicly available at \url{https://physionet.org/content/mimiciv/2.2/}. Access requires completion of the required training course and acceptance of the data use agreement.

\textbf{Materials availability}  
Not applicable.

\textbf{Author contributions}

 Li Sun led manuscript writing and editing and model validation. Junyi Fan conducted experiments, model implementation, and data analysis. Shuheng Chen and Yong Si contributed to data analysis and model validation.  Minoo Ahmadi contributed to feature analysis and data preprocessing. Elham Pishgar provided methodology expertise and manuscript review. Kamiar Alaei contributed to validation and manuscript review. Maryam Pishgar provided supervision, conceptualization, and project oversight. All authors reviewed and approved the final manuscript.





\bibliography{sn-bibliography.bib}

\end{document}